\documentclass[fleqn,10pt]{wlpeerj} 

\usepackage{hyperref}
\usepackage{gensymb}
\usepackage{afterpage}
\usepackage{pdflscape}
\usepackage{tabulary}

\title{skeletracks: automatic separation of overlapping fission tracks
in apatite and muscovite using image processing}

\author[1, 2]{Alexandre Fioravante de Siqueira}
\author[1]{Wagner Massayuki Nakasuga}
\author[1]{Sandro Guedes}
\affil[1]{Departamento de Raios Cósmicos e Cronologia, IFGW, University
          of Campinas, Brazil}
\affil[2]{Institut für Geologie, TU Bergakademie Freiberg, Germany}
\corrauthor[1]{Alexandre Fioravante de Siqueira}{afdesiqueira@gmail.com}

\begin{abstract}
One of the major difficulties of automatic track counting using
photomicrographs is separating overlapped tracks. We address this issue
combining image processing algorithms such as skeletonization, and we
test our algorithm with several binarization techniques. The counting
algorithm was successfully applied to determine the efficiency factor
GQR, necessary for standardless fission-track dating, involving counting
induced tracks in apatite and muscovite with superficial densities of
about $6 \times 10^5$ tracks/cm${}^2$.
\end{abstract}

\begin{document}

\flushbottom
\maketitle
\thispagestyle{empty}

\section*{Introduction}

The fission track dating (FTD) is based on the spontaneous fission of
${}^{238}$U, an impurity in natural minerals such as apatite and zircon
\citep{WAGNER1992}. The fission process releases two fragments. They trigger
the displacement of atoms, leading to structural net alterations called
latent tracks. After convenient etching, channels are formed along the
latent track trajectory and become visible under an optical microscope.
These channels are referred to as tracks, and their number is used to
calculate the fission-track age. Tracks are counted at the microscope or
in photomicrographs captured using a camera coupled to a microscope, in
a time expensive process. Besides, track counting efficiency is observer
dependent; to keep it constant, a trained observer must keep full attention
during the several hours taken to analyze a sample for fission-track
dating. The observer also must perform routine recalibration to maintain
the same counting efficiency over time.

Algorithms for automatic track processing and counting in images from
natural minerals have been proposed (e.g. \cite{WADATSUMI1990, PETFORD1992,
GLEADOW2009, DESIQUEIRA2014, DONELICK1999}). When associated with automatic
systems for capturing photomicrographs, such algorithms have the potential
to increase dating speed. However, these solutions still demand the counting
results to be reviewed, and often adjusted, by the observer, being more
time consuming than manual counting \citep{YASUDA2005, ENKELMANN2012}. The
major challenges to automatic track counting are detecting overlapping
tracks, distinguishing tracks and material defects (e.g. surface scratches
due to polishing), and identifying small tracks and defects of comparable
size in the background of photomicrographs \citep{GLEADOW2009}, which are
straightforward tasks for experienced observers.

In this study we address the issue of automatically separating and counting
overlapping tracks in apatite and muscovite photomicrographs, combining
image processing algorithms. These techniques include the skeletonization
algorithm, which was already proposed for separating overlapping tracks
in a preliminary study \citep{LIPPOLD2007}. The solution presented here
does not exclude previous developments. Instead, it could be combined to
other algorithms, and improve the speed and reliability of track counting.
As a proof of concept, we apply the resulting algorithm to the determination
of the efficiency factor GQR, fundamental for the standardless fission-track
dating \citep{DANHARA2013, JONCKHEERE2003}.

\section*{Material and methods}

In this section we present the photomicrograph test set used in this study,
along with the proposed methodology to separate tracks. It consists in:

\begin{enumerate}
    \item Reading and filtering the input image.
    \item Binarizing the filtered image.
    \item Separating and skeletonizing each region containing candidate
    tracks in the binary image.
    \item Characterizing each pixel in the skeletons according to their
    neighbors.
    \item Classifying tracks in regions, based on the route and the
    Euclidean distance.
\end{enumerate}

\subsection*{Apatite and muscovite photomicrographs}

To illustrate and test the methodology presented here, we used photomicrographs
of co-irradiated natural apatite samples from Durango, Mexico, and muscovite
mica\footnote{These photomicrographs are contained in the folder
\texttt{orig\_figures}, available in the Supplementary Material.} (Figure
\ref{fig:test_images}). To calculate GQR, tracks in both cases are
generated by neutron-induced fission of ${}^{235}$U. This process will be further
detailed in section \ref{sec:calc_gqr}. Durango is a yellowish fluorapatite,
found as well-formed crystals in the Cerro de Mercado iron mine (Durango, Mexico).
Its age, $31.44 \pm 0.18$ Ma \citep{MCDOWELL2005}, and chemical compositions
\citep{DONELICK1999} are well constrained. Durango apatite is widely used
as age standard for (U-Th)/He and fission-track thermochronology. This
sample is largely used also for methodological studies.

\begin{figure*}[htb]  
    \centering
    \includegraphics[width=0.9\textwidth]{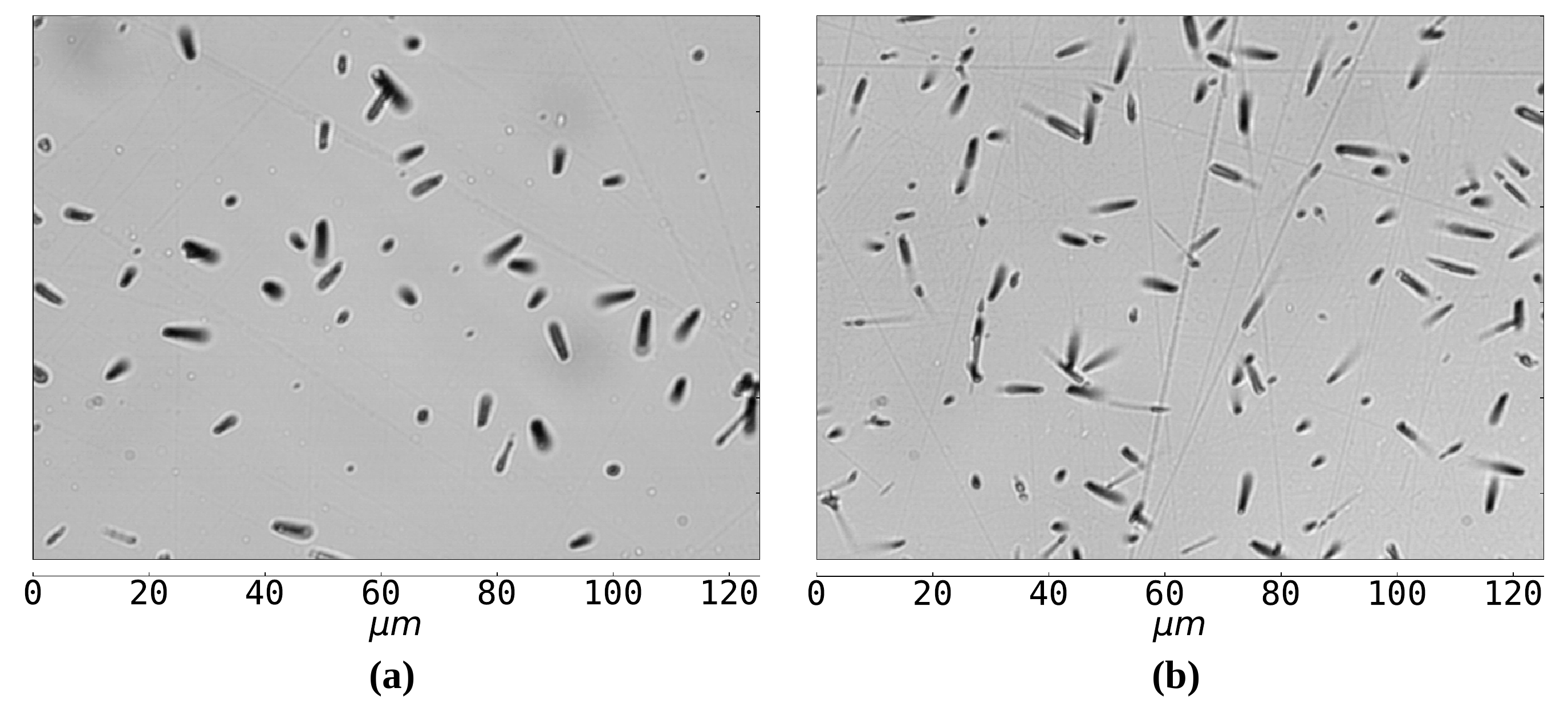}
    \caption{Photomicrographs from the test dataset, presenting fission
    tracks in (a) muscovite mica and (b) apatite samples.}
    \label{fig:test_images}
\end{figure*}

\subsection*{Image filtering}

We used median filters for smoothing the input images before further
processing. After Tukey \citep{TUKEY1971} and others suggested the use of median
filters for signal smoothing, Pratt and Frieden applied these filters for
image processing \citep{PRATT1975, FRIEDEN1976}. The median filtering of an input image
when using an $k \times l$ window ($k$, $l$ being odd integers) is an image
equal to the median of the gray levels of the pixels in a $k \times l$ window
centered at each pixel in the input image \citep{HUANG1979}.

To apply the median filters in the input images, we used the implementation
available in scipy's ndimage \citep{JONES2001}. The filtering window used has
size $7 \times 7$. This smoothing could make small tracks to fade; however,
small defects in the surface of the material would also fade, thus not
being counted as tracks in further processing.

\subsection*{Image binarization}

Photomicrographs captured from Durango apatite mounts were binarized using
several algorithms:

\begin{itemize}
    \item Otsu\citep{OTSU1979}: calculates the optimal threshold based on
    the minimal weighted sum of within-class variances chosen from pixels
    of regions of interest (ROI) and the background.
    \item Yen\citep{YEN1995}: calculates the threshold based on a maximum
    correlation criterion, which uses a cost function. It is a computationally
    efficient alternative to entropy measures.
    \item Li\citep{LI1998}: selects a threshold that minimizes the cross entropy
    between the original and thresholded images.
    \item ISODATA\citep{RIDLER1978}: chooses the threshold using iterations. At
    iteration $n$, a new threshold $T_n$ is calculated using the mean average
    of the classes ``regions of interest'' and ``background''. The process
    is repeated until $T_{n}-T_{n-1}$ becomes sufficiently small. ISODATA
    always converges when applied for two classes \citep{VELASCO1980}.
    \item MLSS\citep{DESIQUEIRA2014}: the user chooses a threshold based
    on a list of results.
\end{itemize}

After binarizing the photomicrographs, we used three tools for enhancing
the result:

\begin{itemize}
    \item\textbf{Excluding small regions:} erasing small regions from the binary
    image avoid small scratches on the surface to be identified as tracks.
    \item\textbf{Filling regions:} tracks may have different gray levels within
    its extent. Filling closes the holes inside these tracks, when they
    are not separated completely by binarization; this avoid finding more
    tracks than desired when using skeletonization.
    \item\textbf{Clearing lower and right borders:} tracks crossing the image
    borders are shared by two counting areas. We use the ``lower right
    corner'' method to avoid counting these tracks twice, which means
    that objects touching the bottom and right edges are not counted.
    Counting tracks over many images reduce this kind of bias \citep{RUSS2011}.
\end{itemize}

\subsection*{Separating and skeletonizing regions}

\begin{figure}[htb]  
    \centering
    \includegraphics[width=0.85\textwidth]{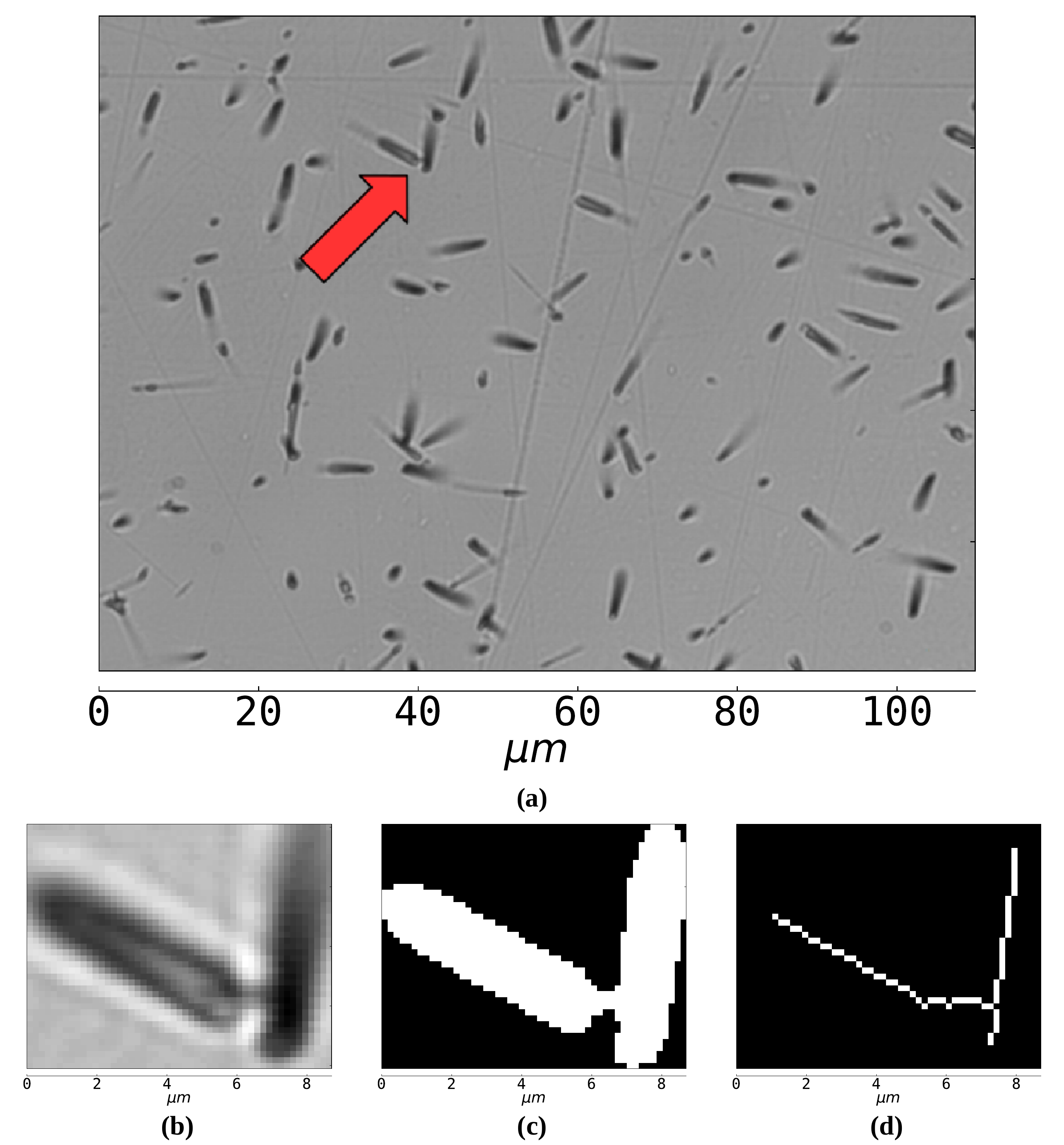}
    \caption{Binarizing and skeletonizing a region highlighted in an input
    photomicrograph. (a) Input photomicrograph containing the example
    region (red arrow). (b) Example region separated from the input
    photomicrograph. (c) Binarizing the region in (b) using the ISODATA
    algorithm (threshold: 133). (d) Skeletonizing the binary region in
    (c). Colormap: gray.}
    \label{fig:img_binskel}
\end{figure}

After binarizing the input photomicrograph, we separate each distinct
region in the binary image. Then, we reduce each region to a single pixel
line using skeletonization \citep{LEE1994} (Figure \ref{fig:img_binskel}).
This process makes it easier to find descriptors for overlapping tracks,
as shown next.

\subsection*{Characterizing pixels}

\begin{figure*}[htb]  
    \centering
    \includegraphics[width=1\textwidth]{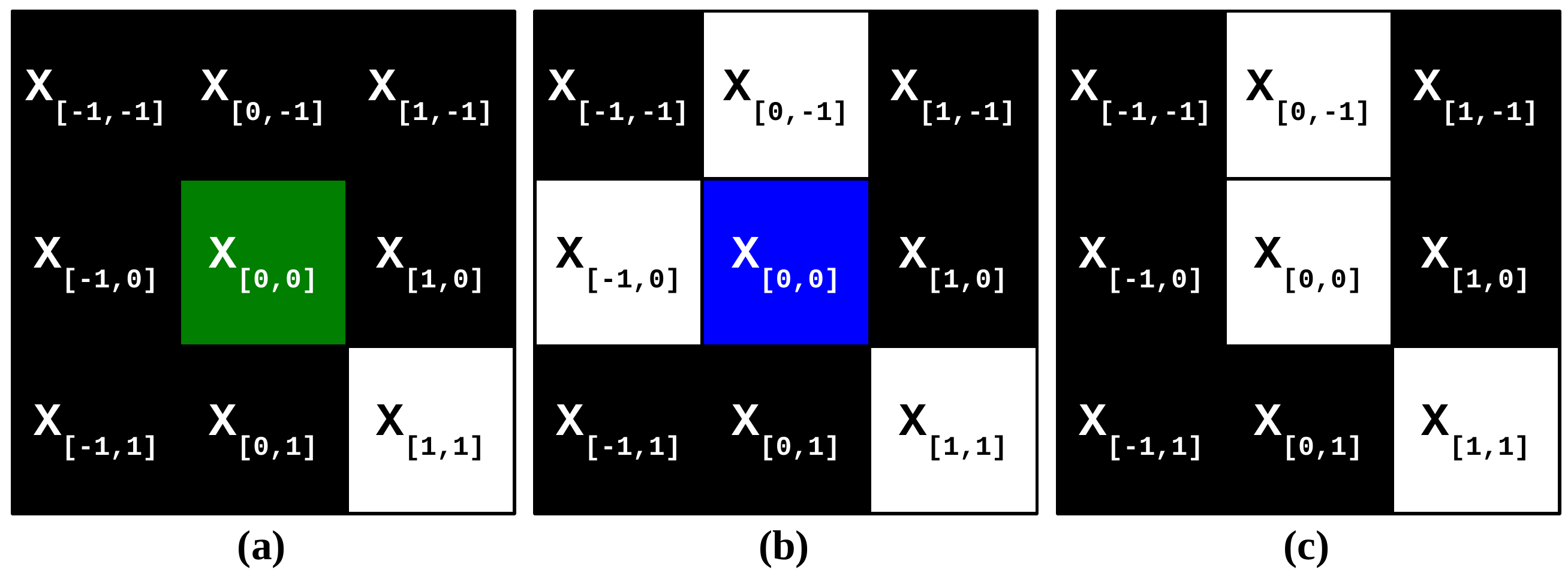}
    \caption{The 8-neighborhood characterized by a central pixel
    $X_{[0, 0]}$ and its neighbors. In this method, (a) $X_{[0, 0]}$
    represents a extremity pixel (only one neighbor, $X_{[1, 1]}$); (b)
    $X_{[0, 0]}$ represents a intersection pixel (three neighbors;
    $X_{[0, -1]}$, $X_{[-1, 0]}$ and $X_{[1, 1]}$); (c) $X_{[0, 0]}$
    represents a pixel in the common case (two neighbors; $X_{[0, -1]}$
    and $X_{[1, 1]}$).}
    \label{fig:px_cat}
\end{figure*}

A major challenge of automatic track counting is distinguishing individual
tracks in a cluster of overlapping tracks. Identifying the points where
tracks intercept each other and where tracks end (their extremities) is
a key step to distinguish cluster geometry. Using the skeletonized regions,
intersections and extremities of tracks can be defined as two pixel sets:

\begin{itemize}
    \item\textbf{Extremity pixels} have only one neighbor in the 8-pixel
    neighborhood (Figure \ref{fig:px_cat}(a)), thus indicating possible
    extremities of a track.
    \item\textbf{Intersection pixels} have more than two neighbors in the
    8-pixel neighborhood (Figure \ref{fig:px_cat}(b)), indicating a possible
    overlapping between two or more tracks.
\end{itemize}

There is also the common case, where a pixel has two neighbors belonging
to the same region (Figure \ref{fig:px_cat}(c)).

The simplest kind of track cluster contains only two tracks, which present
three extremity and one intersection pixel (Figure 4). Considering these
pixels, when the region does not have intersection pixels it represents
only one track. When the region contains intersection pixels, it represents
more than one track.

\begin{figure*}[htb]  
    \centering
    \includegraphics[width=1\textwidth]{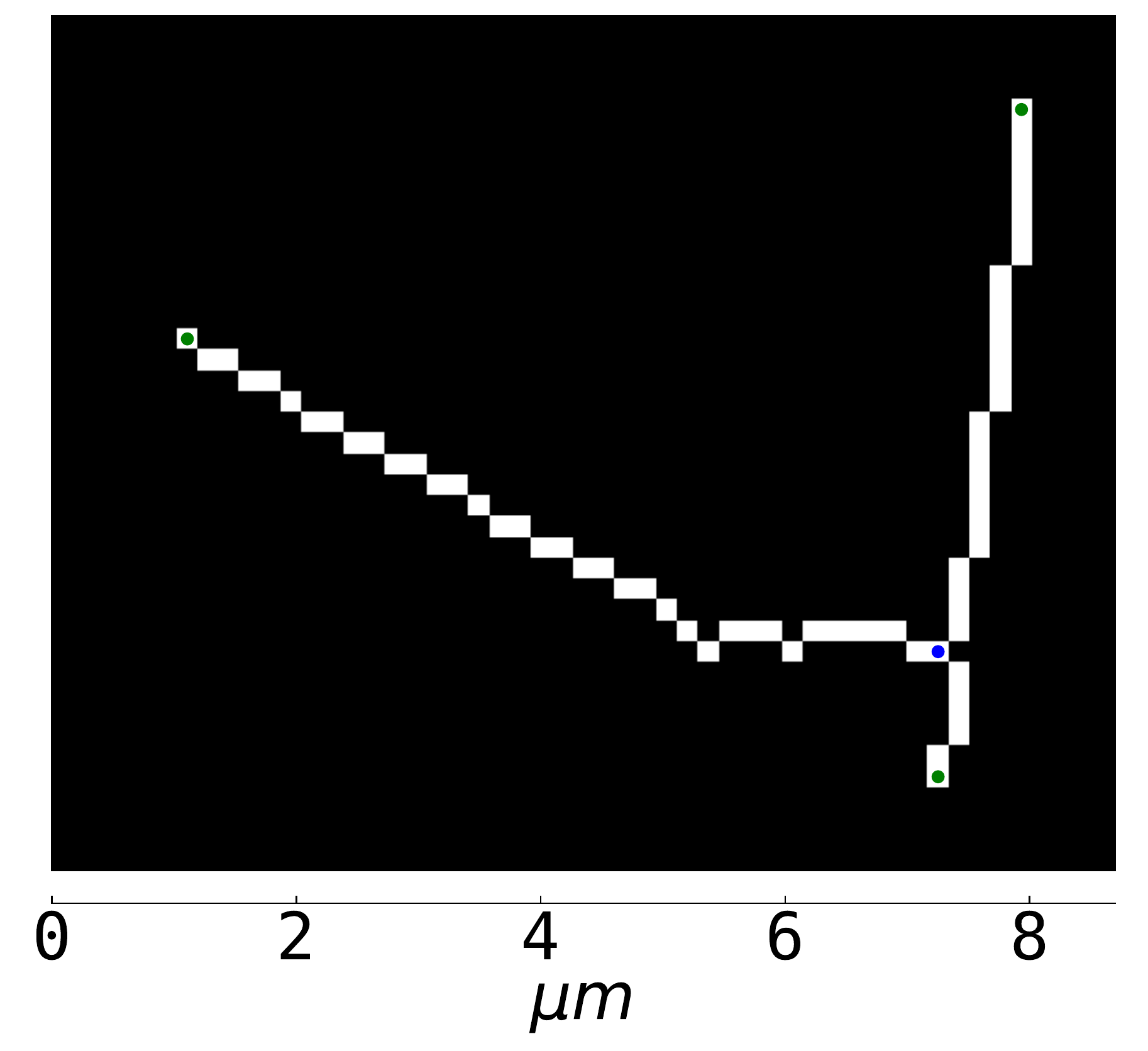}
    \caption{Labeling the extremity (green dots) and intersection (blue
    dot) pixels in the skeletonized region (Figure
    \ref{fig:img_binskel}(d)).}
    \label{fig:px_exextint}
\end{figure*}

\subsection*{Classifying track candidates in regions}

\begin{figure*}[htb]  
    \centering
    \includegraphics[width=1\textwidth]{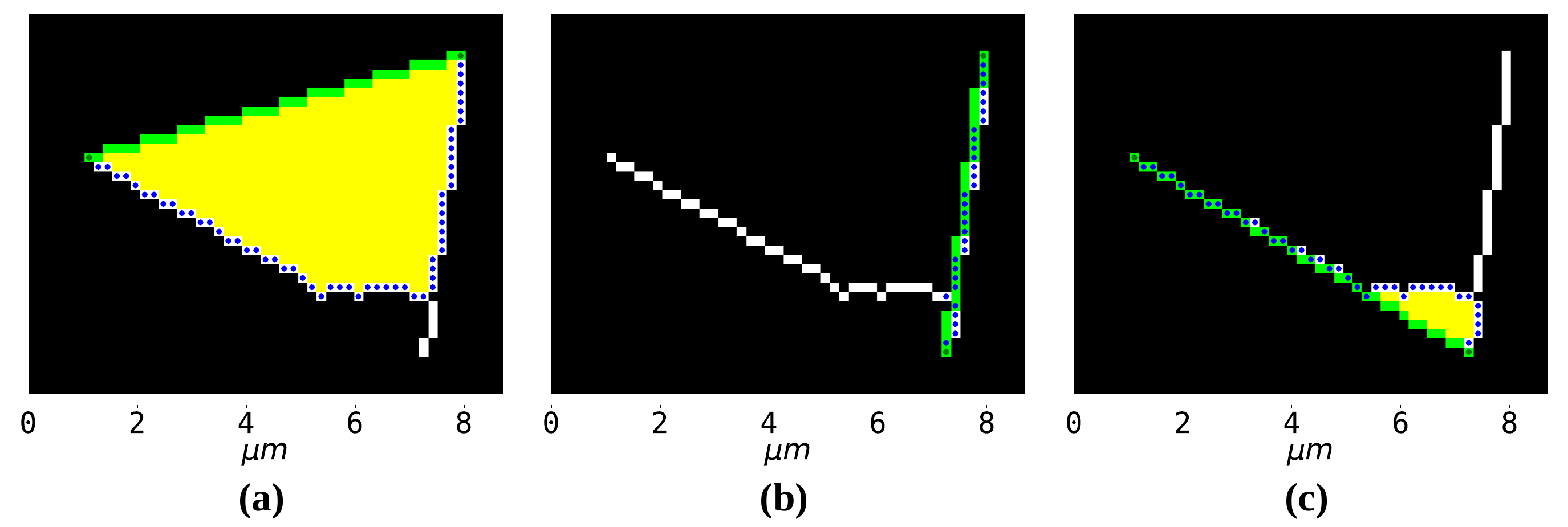}
    \caption{Choosing track candidates in the region presented in Figure
    \ref{fig:img_binskel}(b), obtaining extremity points two by two. Green
    pixels: Euclidean distance. Blue dots: route between the two extremity
    points. Yellow pixels: inner area of the region formed by Euclidean
    distance and route.}
    \label{fig:all_trackcand}
\end{figure*}

After obtaining the extremity and intersection pixels from a region, we
choose the track candidates:

\begin{enumerate}
    \item Extremity pixels are grouped in pairs. If there are only two
    extremity pixels, the region represents only one track.
    \item We calculate the Dijkstra's minimum cost path \citep{DIJKSTRA1959}
    passing by an intersection pixel, referred here as route, and the
    Euclidean distance between the two chosen pixels. The union of these
    curves form a region in the binary image.
    \item The inner area from the region generated by the route and the
    Euclidean distance is given by the pixels within this region. The
    first track candidate is defined as the Euclidean distance between
    the pair of pixels yielding the smallest inner area (Figure
    \ref{fig:all_trackcand}).
\end{enumerate}

\begin{figure*}[htb]  
    \centering
    \includegraphics[width=1\textwidth]{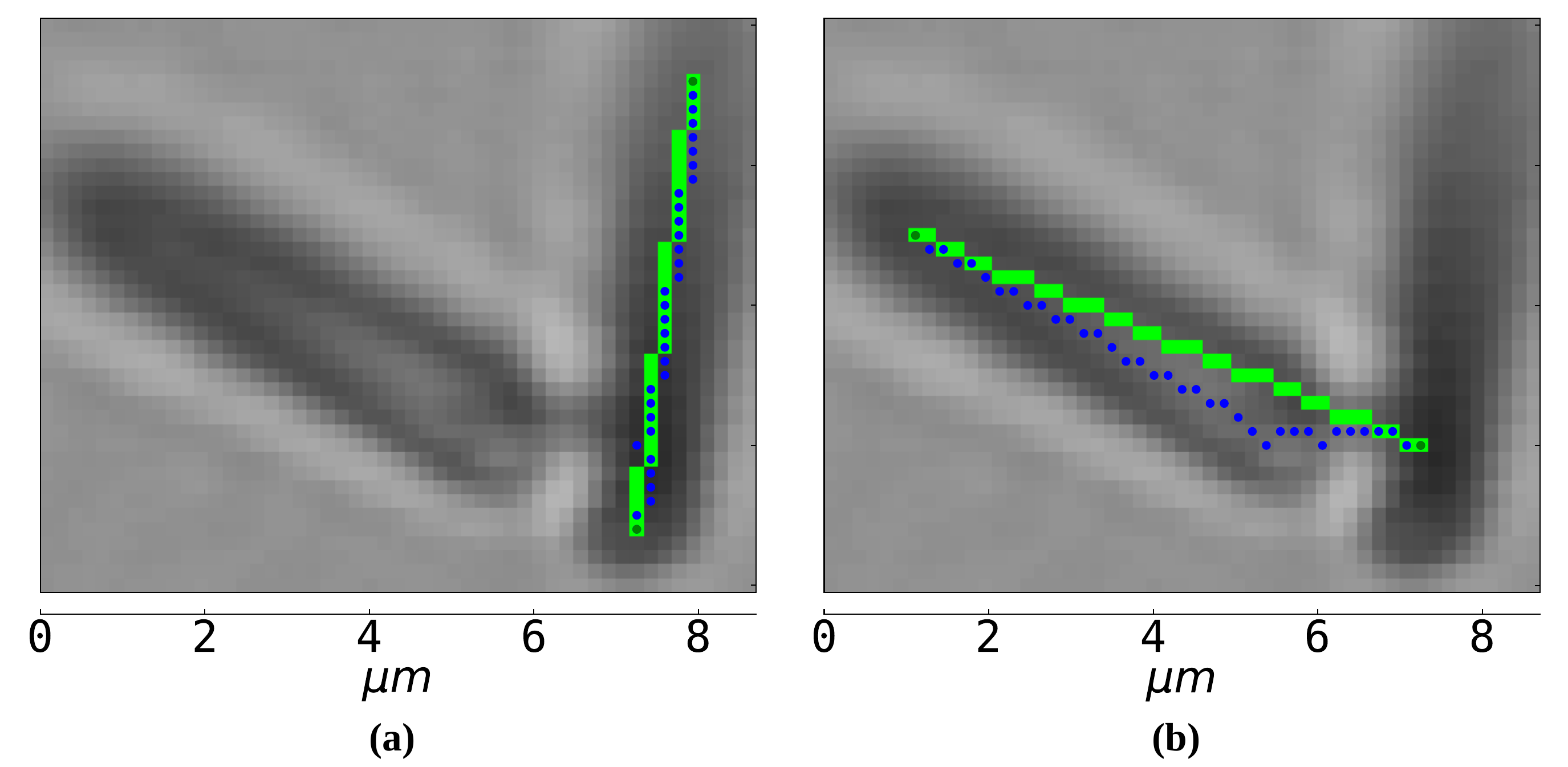}
    \caption{Track candidates chosen by the algorithm for the region in
    Figure \ref{fig:img_binskel}(b). Green dots: extremity pixels. Green
    line: Euclidean distance between extremity pixels. Blue dots: route
    between extremity pixels.}
    \label{fig:final_trackcand}
\end{figure*}

After identifying the track candidate, its extremity pixels are excluded
and the process starts again. The process continues until every extremity
pixel in the skeleton has a pair (Figure \ref{fig:final_trackcand}). If
the number of extremity pixels in the skeleton is odd, the last extremity
pixel left will unite with the closest intersection pixel (Figure
\ref{fig:final_trackcand}(b)).

\begin{figure*}[htb]  
    \centering
    \includegraphics[width=1\textwidth]{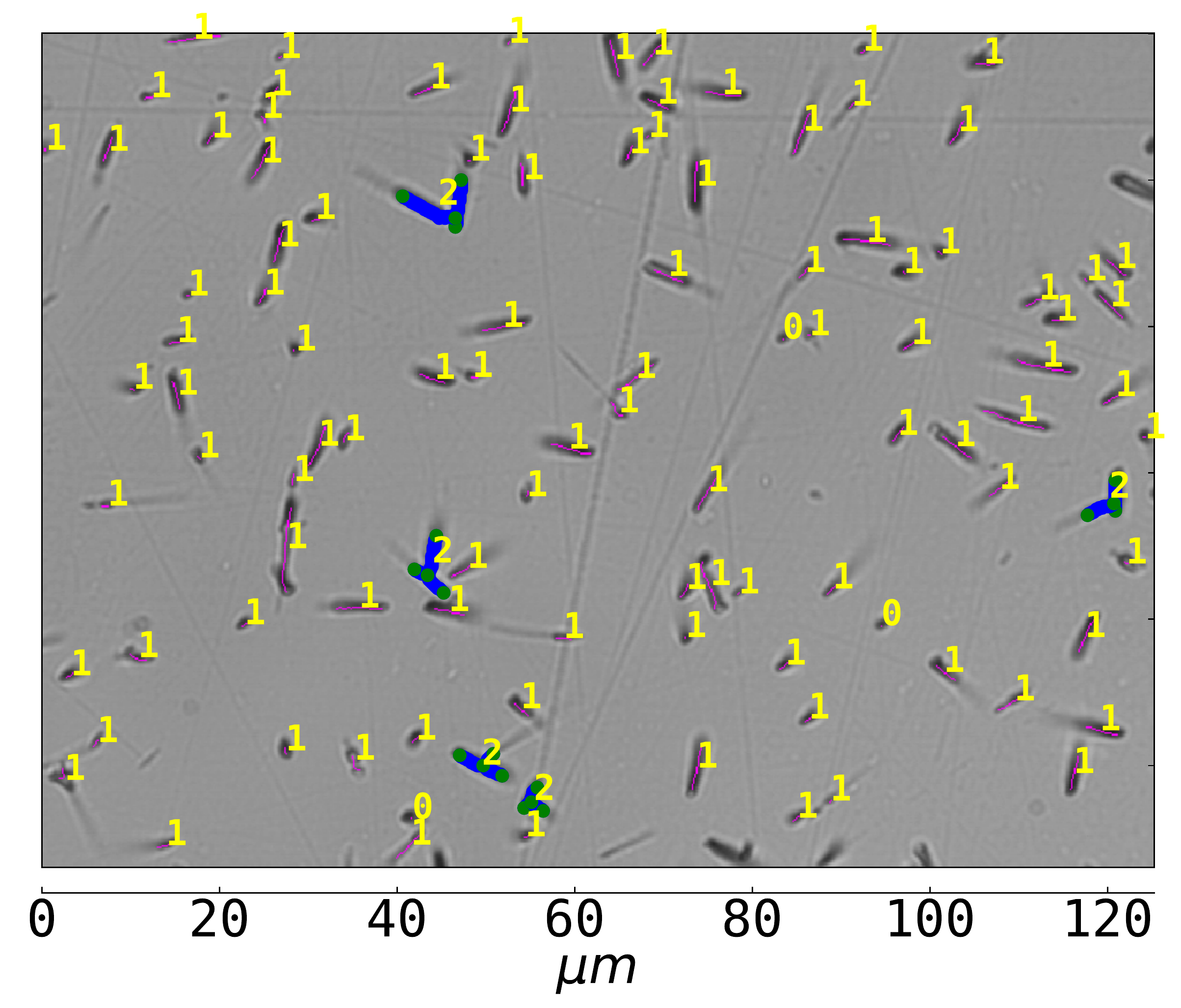}
    \caption{Visual representation of each track labeled by the segmentation
    algorithm, when using the ISODATA binarization (threshold: ~0.475).
    The numbers show how many tracks were counted in each region. Magenta
    lines: regions representing only one track. Green dots: extremity pixels.
    Green lines: Euclidean distance between extremity pixels. Blue paths:
    route between extremity pixels. }
    \label{fig:final_fig}
\end{figure*}

All binary regions containing single tracks or track clusters are processed.
Then, the algorithm returns the number of tracks contained in the input
photomicrograph and the visual representation of each track labeled by
the algorithm (Figure \ref{fig:final_fig}).

\subsection*{Calculating GQR for Standardless Fission-Track Dating}
\label{sec:calc_gqr}

The External Detector Method (EDM) is the dating approach chosen in most
fission-track studies. Apatite grains containing fossil fission tracks
are mounted in epoxy resin, grounded, polished (to reveal an internal
plan surface) and etched. A muscovite mica sheet is coupled to the apatite
mount and the set is irradiated with neutrons in a nuclear reactor. The
mica sheet is also etched after irradiation. The densities of tracks in
the apatite grains, $\rho_S$, and in the co-irradiated mica, $\rho_I$,
are determined. The standardless fission-track age equation can be written
as \citep{JONCKHEERE2003}:

\begin{equation}
t = \frac{1}{\lambda} \ln\left ( 1+GQR \frac{\rho_S}{\rho_I} \frac{\lambda R_U}{\lambda_f C_{238}}\right )
\label{eq:ft_age}
\end{equation}

In Eq. \ref{eq:ft_age}, $\lambda$ is the total decay constant, $\lambda_f$ is
the spontaneous fission decay constant, and $C_{238} $ is the isotopic
concentration of ${}^{238} U$. $R_U$, the number of induced fission reactions per
uranium atom, characterizes the neutron irradiation and is determined
experimentally (for instance in \citep{IUNES2002, SOARES2014}). The decay
constant $\lambda_f$ and the direct quantification of neutron fluence,
which once were controversial (e.g. \cite{HADLER2003}) have
been long resolved. Fission-track determinations in the decade of 2000
\citep{GUEDES2000, GUEDES2003, SUZUKI2005, YOSHIOKA2005} confirmed the
value recommended by the International Union of Pure and Applied Chemistry
(IUPAC), $\lambda_f = (8.5 \pm 0.1) \times 10^{-17} a^{-1}$
\citep{HOLDEN2000}. Extensive calibration of neutron quantification has
also been carried out \citep{DECORTE1991, VANDENHAUTE1988, CURVO2013,
SOARES2014}.

The geometry factor G accounts for the different tracks source volumes
generating tracks in apatite and mica. Apatite tracks are generated by
fragments coming from below and above the exposed surface ($4 \pi$-geometry)
while mica tracks are generated only by fragments coming from the apatite
($2\pi$-geometry). Ideally, $G = 1/2 (2\pi/4\pi)$. Q is a procedural factor
\citep{JONCKHEERE2003} which considers the different etching and observation
efficiencies between apatite and mica. R is the range deficit factor accounting
for differences in the ability of mica and apatite to record etchable tracks
\citep{IWANO1998}. Although it has been shown that these factors can have their
values calculated separately \citep{JONCKHEERE2003}, it is possible to determine
the product GQR experimentally, usually using a standard sample.

To determine the value of GQR, we applied the technique of comparing the
density of induced tracks in an internal apatite surface, $(\rho_{I})_{IS}$,
with the density of induced tracks in the external detector, $(\rho_{I})_{ED}$,
coupled with the mineral during neutron irradiation. It has been shown
(e.g. \cite{JONCKHEERE2003, SOARES2013}) that:

\begin{equation}
GQR = \frac{(\rho_{I})_{ED}}{(\rho_{I})_{IS}}
\label{eq:gqr}
\end{equation}

Three crystals of Durango apatite were mounted in epoxy resin, grounded
and polished. A plan sheet of muscovite mica was coupled with the apatite
mount and this set was irradiated with neutrons in the IEA R1 reactor at
the IPEN/CNEN, São Paulo, Brazil. Neutron fluence was monitored with a CN1
glass. The value of $R_U = (3.2 \pm 0.1) \times 10^{-8}$ fissions per
uranium atom (equivalent to a neutron fluence of approximately
$7.4 \times 10^{15}$ neutrons/$cm^2$) was adopted using the calibration
proposed by \citep{SOARES2014}. After irradiation, the apatite mount was
grounded and polished again to reveal an internal surface, where tracks
were etched in 1.1 molar nitric acid for $50 s$ at $20\degree C$. Muscovite
mica was etched in 48\% HF for 120 minutes at $15\degree C$. Thirty apatite
and 49 muscovite photomicrographs were captured with a nominal magnification
of $500 \times$ at a Zeiss Axioplan II optical microscope. The track densities
were determined by manual counting, and using the counting algorithm
proposed in this study.

\subsection*{License and reusability}

The algorithms and functions implemented in this study were built using
the Python packages Numpy \citep{OLIPHANT2006, VANDERWALT2011}, Scipy \citep{JONES2001},
Matplotlib \citep{HUNTER2007}, scikit-image \citep{VANDERWALT2014},
Jansen-MIDAS \citep{DESIQUEIRA2018}, among others. All code published
with this paper is available at
\url{https://github.com/alexandrejaguar/publications/tree/master/2018/skeletracks}.
It is written in Python 3 \citep{VANROSSUM1995}, and is available under
the GNU GPL v3.0 license. All photomicrographs and figures distributed
with this paper are available under the CC-BY 2.0 license.

\section*{Results}

\begin{figure}[htb]  
    \centering
    \includegraphics[width=1\textwidth]{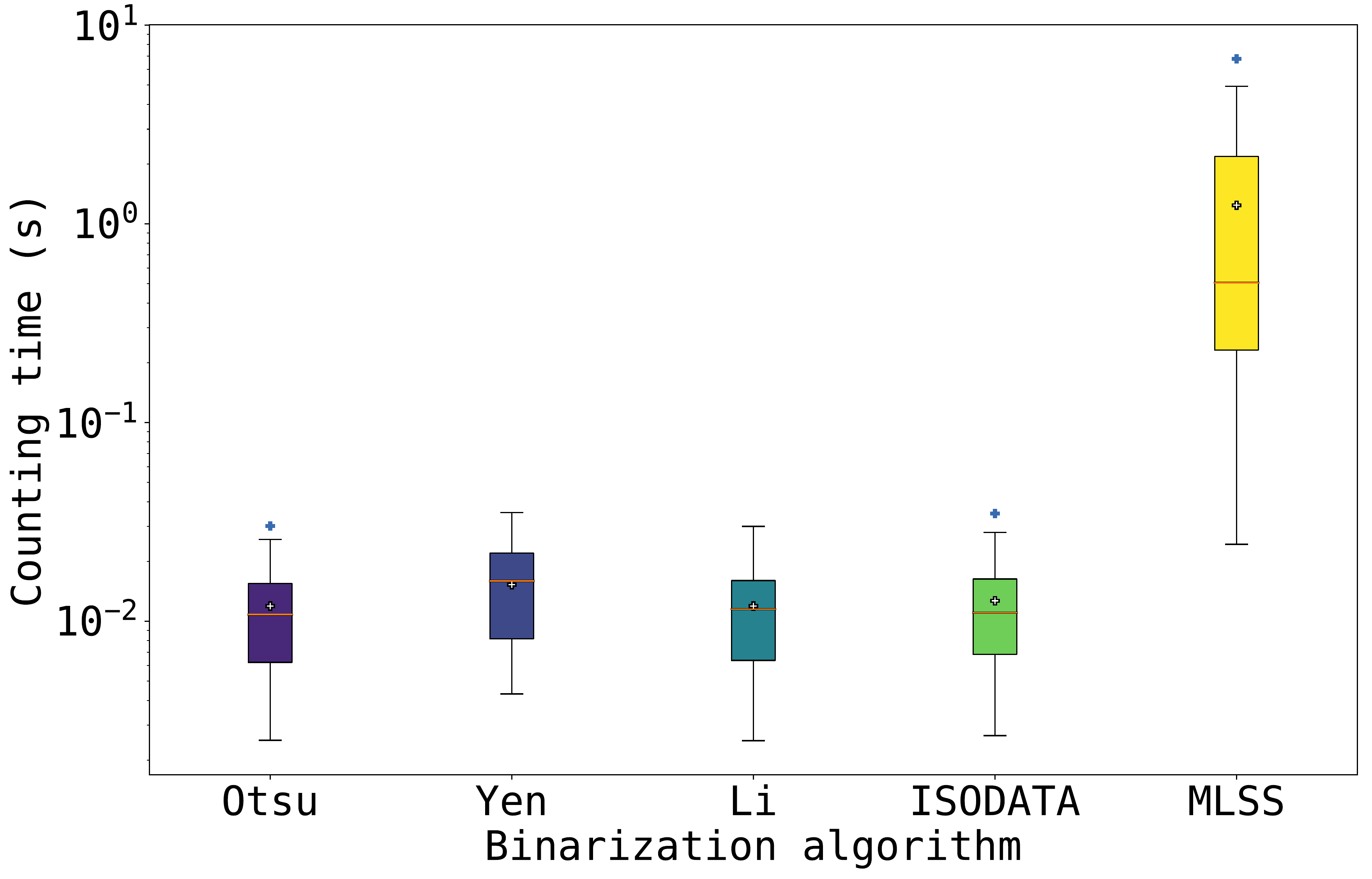}
    \caption{Track counting time for images in the test dataset, according
    to each binarization algorithm. Counting times are usually small
    (around 10${}^{-2}$ s), except for MLSS, which employs wavelet decompositions
    in it \citep{DESIQUEIRA2014}, being more time demanding than the other
    binarizations. Another fact would be that MLSS adds artificial regions
    to the binary image; then, further track counting is impaired when
    using this binarization.}
    \label{fig:count_times}
\end{figure}

We counted all images in the test dataset using Otsu, Yen, Li, ISODATA,
and MLSS binarizations. Counting times are smaller than $10^{-1}\, s$ when using most binarization
algorithms (Figure \ref{fig:count_times}), except for MLSS: it employs
wavelet decompositions for segmenting the input image \citep{DESIQUEIRA2014},
being more time demanding than the other algorithms. Also, further track
counting could be impaired when using MLSS, since it adds artificial regions
to the binary image as discussed in section \ref{sec:discussion}.

\begin{figure*}[htb]  
    \centering
    \includegraphics[width=1\textwidth]{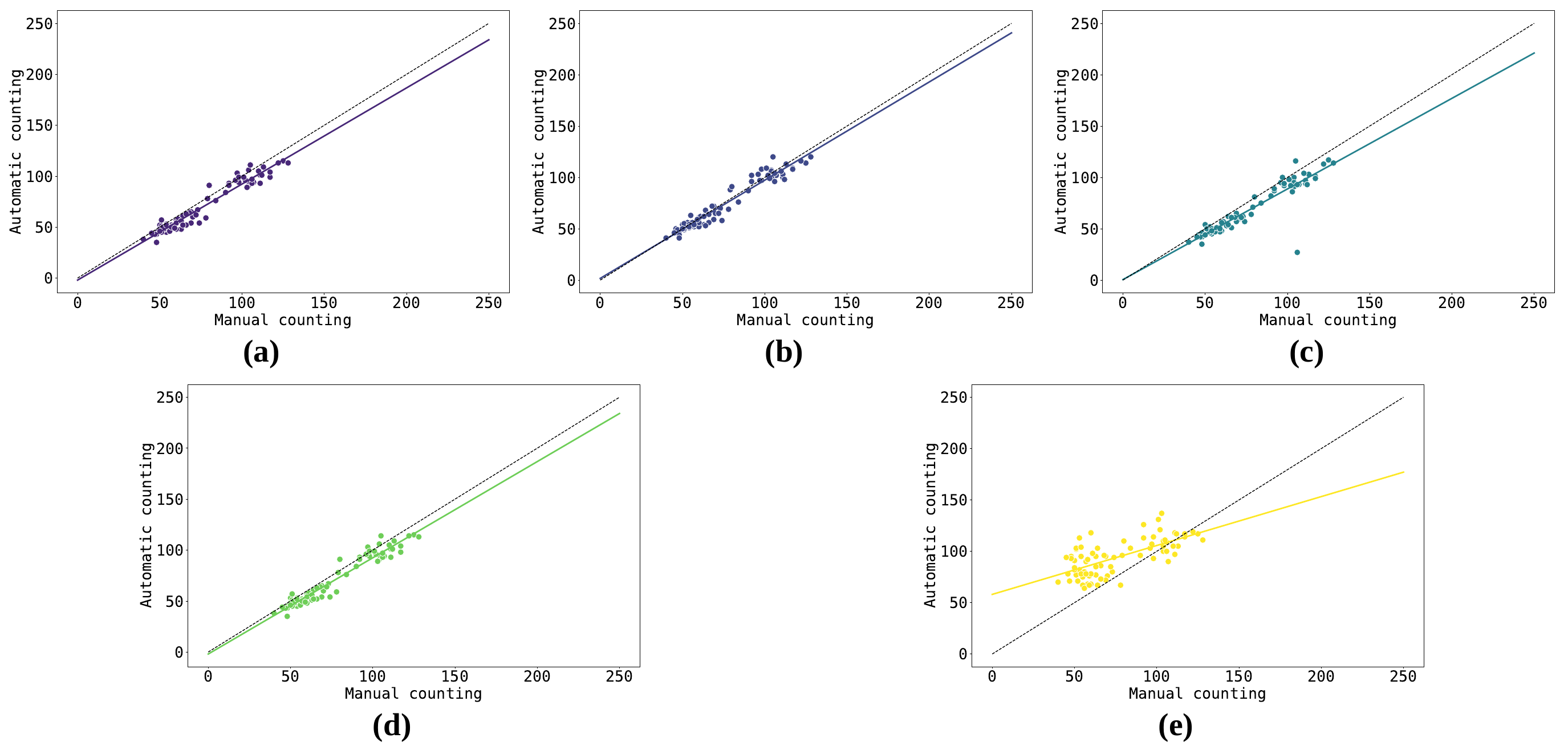}
    \caption{Comparison between manual and automatic counting with the
    proposed method in the test photomicrographs. The colored lines
    represent the data linear regression. Binarizations used:
    (a) Otsu, (b) Yen, (c) Li, (d) ISODATA, and (e) MLSS. Dashed line:
    1:1 line.}
    \label{fig:manauto_count}
\end{figure*}

Different binarization algorithms may yield different number of track
candidates, and therefore different track counting, depending on the
complexity of the input region (Figure \ref{fig:manauto_count}). In general,
the proposed algorithm can recognize the tracks in the binary regions.

Using the proposed algorithm, we counted the fission tracks contained in
the test photomicrographs obtained from the Durango apatite and co-irradiated
mica. Additionally, one of the authors (W. N.) manually performed the same
measurements (Table \ref{tab:count_data}). Apatite and mica counts have
been pooled to calculate $\rho_{IS}$ and $\rho_{ED}$. The uncertainty presented
is one Poisson standard deviation for the densities and the usual propagation
of uncertainties for GQR:

\begin{equation}
\frac{\sigma_{GQR}}{GQR} = \sqrt{\frac{1}{N_{ED}} + \frac{1}{N_{IS}}}
\end{equation}

\centering
\begin{table*}[h]  
    {\tiny
        \begin{tabulary}{1.8\textwidth}{CCCCCCCCCCC}
            \textbf{Counting detail} & \textbf{$N_{ED}$} & \textbf{$N_{ED}/image\,(\pm 1 \sigma)$} & \textbf{$D_{ED}$} & \textbf{$(p-value)_{ED}$} & \textbf{$N_{IS}$} & \textbf{$N_{IS}/image\,(\pm 1 \sigma$)} & \textbf{$D_{IS}$} & \textbf{$(p-value)_{IS}$} & \textbf{$\rho_{ED} \pm 1 \sigma\,(\times 10^5\,tracks/{cm}^2$)} & \textbf{$GQR$} \\
            \hline
            \textbf{Otsu} & 2133 & $45.5 \pm 0.9$ & 0.10 & 0.63 & 2279 & $76.0 \pm 1.6$ & 0.11 & 0.82 & $3.70 \pm 0.08$ & $0.57 \pm 0.02$ \\
            \textbf{Yen} & 2407 & $49.1 \pm 1.0$ & 0.12 & 0.42 & 2484 & $82.8 \pm 1.7$ & 0.14 & 0.59 & $4.17 \pm 0.09$ & $0.59 \pm 0.02$ \\
            \textbf{Li} & 2079 & $42.4 \pm 0.9$ & 0.16 & 0.17 & 2150 & $71.7 \pm 1.6$ & 0.20 & 0.14 & $3.60 \pm 0.08$ & $0.59 \pm 0.02$ \\
            \textbf{ISODATA} & 2136 & $43.6 \pm 0.9$ & 0.11 & 0.61 & 2282 & $76.1 \pm 1.6$ & 0.10 & 0.90 & $3.70 \pm 0.08$ & $0.57 \pm 0.02$ \\
            \textbf{MLSS} & 5583 & $113.9 \pm 1.5$ & 0.23 & 0.01 & 3346 & $111.5 \pm 1.9$ & 0.30 & 0.01 & $9.68 \pm 0.13$ & $1.02 \pm 0.02$ \\
            \textbf{Manual} & 2838 & $57.9 \pm 1.1$ & 0.10 & 0.67 & 3114 & $103.8 \pm 1.9$ & 0.09 & 0.93 & $4.91 \pm 0.09$ & $0.56 \pm 0.01$ \\
            \hline
        \end{tabulary}
    }
    \caption{Data used for calculating GQR. Tracks were counted in 49 mica
             and 30 apatite images. Image area: $3.58\,cm^2$. Counting
             detail: binarization algorithm combined with automatic counting,
             or manual counting; $N_{ED(IS)}$: number of tracks counted
             in mica (apatite); $\rho_{ED(IS)}$: induced fission track
             density in mica (apatite).}
    \label{tab:count_data}
\end{table*}

To verify whether the counting distributions were drawn from the Poisson
distribution, we used the one-sample, two-sided, Kolmogorov-Smirnov (KS)
test, implemented in the R package ``stats'' \citep{RCORETEAM2018} (Table
\ref{tab:count_data}). The KS test compares cumulative distribution
functions and its statistics, D, is defined as the maximum horizontal
distance between them. We adopted the confidence level of $\alpha = 0.05$.
In the KS test, higher p-values mean that there is no statistical evidence
for differences between the tested distributions.

\section*{Discussion}
\label{sec:discussion}

We proposed an algorithm for counting tracks in binary images obtained
from mineral photomicrographs. Using this algorithm, we calculated GQR
for a test dataset created to this purpose. Since the binarization algorithms
may return different binary results for certain regions, the method
presented here returns different track counting for the same regions
when using different binary input images (Table \ref{tab:count_data},
Figure \ref{fig:manauto_count}).

All counting was carried out in images with track densities of about
$6 \pm 10^5$ tracks/$cm^2$. The counting algorithm should perform well
in samples with track densities within this magnitude or less. From the
KS test, we did not find evidence that most distributions of number of
tracks per image were not drawn from Poisson distribution ($p-value > 0.05$),
as it should be expected from measurements of processes such as the neutron
induced fission of ${}^{235} U$. Only counting with the MLSS binarization
for both apatite and mica tracks resulted in a KS $p-value < 0.05$.

Except when using MLSS binarization, the GQR values found in this study
with automatic counting, $0.57-0.59 \pm 0.02$, are in the range commonly
found in the literature: $0.51 \pm 0.02$ \citep{GLEADOW1977}, $0.55 \pm 0.02$
\citep{IWANO1998}, $0.56 \pm 0.03 - 0.61 \pm 0.02$ \citep{SOARES2013},
$0.58 \pm 0.02$ \citep{IWANO2018}, $0.61 \pm 0.01$ \citep{ENKELMANN2003}.
The same holds for the manual counting.

\begin{figure*}[htb]  
    \centering
    \includegraphics[width=1\textwidth]{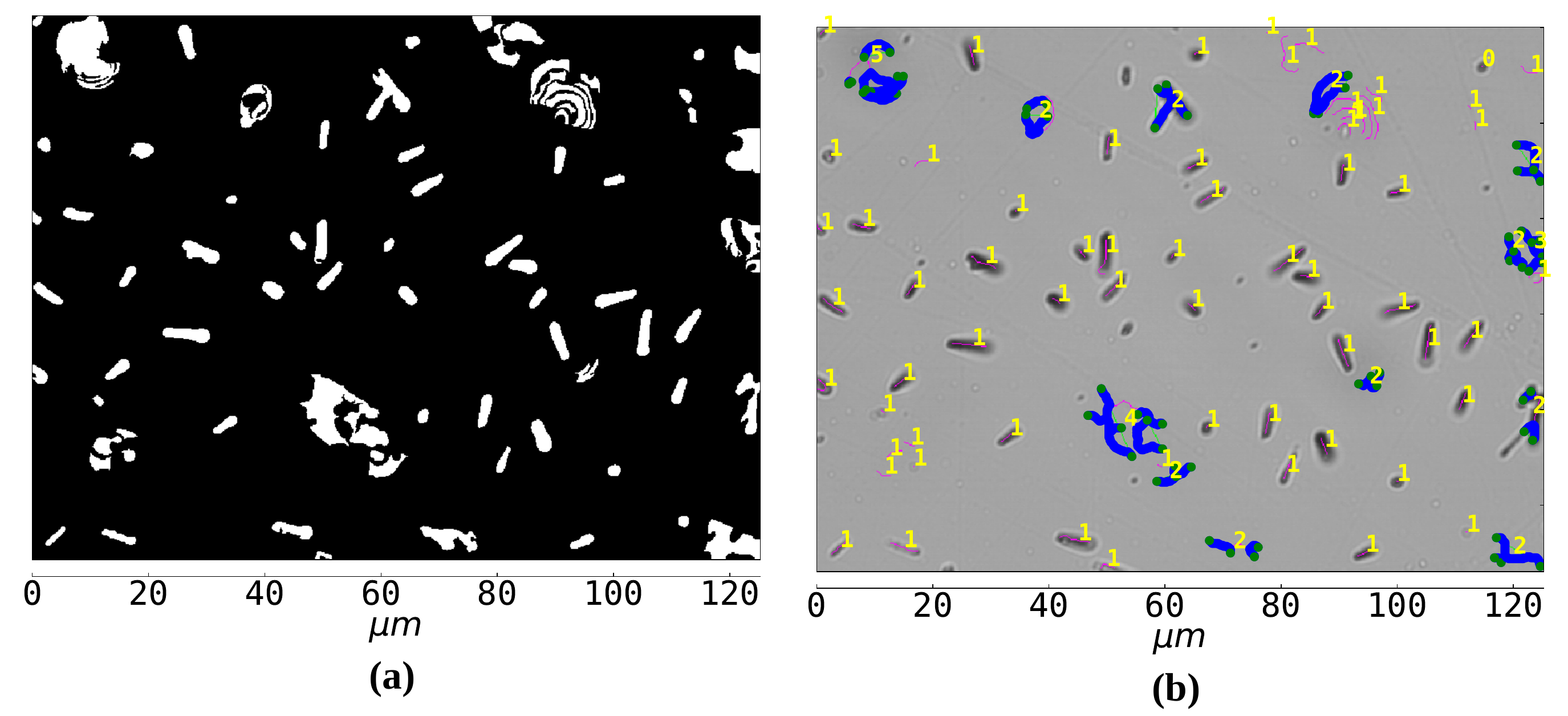}
    \caption{Counting tracks in Figure \ref{fig:test_images}(a). MLSS
    binarization creates artifacts in the resulting binary image, thus
    misleading the track counting algorithm, which counts 115 tracks. (a)
    MLSS binary image obtained from Figure \ref{fig:test_images}, presenting
    the generated artifacts. (b) Results of the automatic counting algorithm.
    Manual counting: 54 tracks. Automatic counting using ISODATA, Li, Otsu,
    and Yen binarizations, respectively: 41, 43, 41, and 44 tracks.}
    \label{fig:mlss_fail}
\end{figure*}

The number of tracks identified as such and counted by the automatic
algorithm is lower than tracks counted manually by the observer, again
with the exception of MLSS binarization. Due to previous processing given
in filtering and binarization, smaller and/or lighter gray tracks are not
counted. The algorithm also does not count tracks in certain cluster
configurations. Although the number of counted tracks is smaller in relation
to manual counting, this feature prevents the count of small artifact
objects like etching figures, avoiding false positives. Once the counting
efficiency remains constant when counting tracks for dating, the age
results should not be affected. On the other hand, MLSS binarization
produces a greater number of counted tracks than manual counting (Figure
\ref{fig:mlss_fail}). In this case, the overidentification of tracks is
due to the counting artifacts (false positives), which compromises dating.

Track counting times for the proposed algorithm are usually small (Figure
\ref{fig:count_times}), except for MLSS. This binarization algorithm employs
wavelet decompositions in it \citep{DESIQUEIRA2014}, being more time demanding
than the other ones. Since MLSS adds artificial regions to the binary image
(Figure \ref{fig:mlss_fail}, Table \ref{tab:count_data}), the further track
counting procedure becomes more time expensive.

Due to its need of external points to recognize track candidates, the
presented algorithm has some limitations. For instance, when the skeletonization
of a certain region results in a single pixel, the algorithm does not have
two external points to process, and will consider the region as having zero
tracks (regions labeled with the number 0 in Figure \ref{fig:final_fig}).

The Euclidean distance of the track candidates may provide a measure of
the track projection in the X axis (green lines in Figure \ref{fig:all_trackcand}).
However, this distance is obtained from the external pixels, defined
according to the skeletonization algorithm (magenta lines in Figures
\ref{fig:final_fig} and \ref{fig:mlss_fail}). Since the skeleton of a region is smaller than the
binary region itself, the attributed distance is slightly smaller than the
actual Euclidean distance of the track extremities. Methods such as Fews\'s
dynamic perimeter positioning \citep{FEWS1992} could emphasize gray levels in
the borders of fission tracks, highlighting them, thus improving the counting
results for overlapped tracks.

\section*{Conclusion}

In this study, we present an algorithm to separate and count overlapping
fission tracks in photomicrographs. This method is based on the definition
of extremity and intersection pixels over the skeleton of a binary region.
Once these pixels are determined, we obtain the inner area of the region
formed by the route between two pixels and their Euclidean distance.
Smallest areas lead to possible track candidates.

Results found for GQR are encouraging since the values found are in the
expected range and false positives are avoided when using most binarization
algorithms. The only exception is the MLSS binarization, which performs
poorly in avoiding false positives, and have higher processing times.

The proposed method is not a definitive solution, but is certainly a step
forward in the automatic counting of tracks. It also can be applied jointly
with other algorithms. The counting algorithm should perform well in samples
with track densities of about $6 \pm 10^5$ tracks/$cm^2$ or less.

\section*{Acknowledgements}

A.F. de Siqueira would like to thank Raymond Jonckheere and Lothar Ratschbacher
for their input on the first ideas of this study. This work is supported by the
São Paulo Research Foundation (FAPESP), grants \# 2014/22922-0 and 2015/24582-4,
and the National Council for Scientific and Technological Development
(CNPq), grant \# 309142/2015-6.


\bibliographystyle{apalike}
\bibliography{afdesiqueira2018_refs}

\end{document}